\documentclass[11pt,twocolumn]{article}

\usepackage[margin=1in]{geometry}
\usepackage{booktabs}
\usepackage{graphicx}
\usepackage{amsmath}
\usepackage{siunitx}
\usepackage[numbers]{natbib}
\usepackage{hyperref}
\usepackage{float}
\usepackage{caption}
\usepackage{subcaption}
\usepackage{placeins}

\title{AMBIT: Augmenting Mobility Baselines with Interpretable Trees}
\author{
Qizhi Wang\\
PingCAP, Data \& AI-Innovation Lab\\
Beijing, China\\
\texttt{qizhi.wang@pingcap.com}
}
\date{}

\begin{document}
\maketitle

\begin{abstract}
Origin--destination (OD) flow prediction remains a core task in GIS and urban analytics, yet practical deployments face two conflicting needs: high accuracy and clear interpretability. This paper develops AMBIT, a gray-box framework that augments physical mobility baselines with interpretable tree models. We begin with a comprehensive audit of classical spatial interaction models on a year-long, hourly NYC taxi OD dataset. The audit shows that most physical models are fragile at this temporal resolution; PPML gravity is the strongest physical baseline, while constrained variants improve when calibrated on full OD margins but remain notably weaker. We then build residual learners on top of physical baselines using gradient-boosted trees and SHAP analysis, demonstrating that (i) physics-grounded residuals approach the accuracy of a strong tree-based predictor while retaining interpretable structure, and (ii) POI-anchored residuals are consistently competitive and most robust under spatial generalization. We provide a reproducible pipeline, rich diagnostics, and spatial error analysis designed for urban decision-making.
\end{abstract}

\section{Introduction}
Urban mobility modeling is foundational for transport planning, infrastructure allocation, and real-time operations. Traditional spatial interaction models (e.g., gravity, radiation, intervening opportunities) are attractive because they are grounded in physical intuitions such as distance decay and mass attraction. However, these models often underperform on modern, high-resolution mobility data. Conversely, deep models achieve high accuracy but are frequently criticized for opaque decision logic and heavy compute requirements. Recent work in GIS is increasingly turning to physics-guided or gravity-inspired learning models, demonstrating that hybrid approaches are both publishable and impactful \citep{gsd_dg_2025,deep_gravity_2021,pag_stan_2025}. At the same time, surveys and critiques emphasize the need for interpretable, efficient models and more realistic evaluation of spatial interaction baselines \citep{mobility_review_2025,sim_limitations_2020,gravity_migration_2022}.

AMBIT addresses this gap by explicitly structuring prediction into a physical baseline plus a learnable residual. The physical baseline encodes distance and attraction (interpretable), while the residual captures deviations due to local urban structure and temporal effects. We evaluate this approach at city scale with hourly OD flows and provide a detailed audit of physical models to clarify what does and does not work at this resolution.

\section{Related work}
\paragraph{Spatial interaction baselines.} Gravity and radiation models remain canonical baselines for OD flows because of their physical interpretation and parameter parsimony \citep{radiation_2012}. However, multiple studies show these baselines can fail to explain temporal dynamics or achieve accurate predictions when evaluated rigorously \citep{sim_limitations_2020,gravity_migration_2022}.

\paragraph{Hybrid and deep OD models.} Recent GIS and remote sensing literature reports gravity-inspired deep networks and physics-guided spatiotemporal attention models \citep{deep_gravity_2021,gsd_dg_2025,pag_stan_2025}. Other works apply GCN/GRU variants and encoder--decoder style OD predictors with strong accuracy but limited interpretability \citep{gcn_gru_2024}. Because OD data are sparse and zero-inflated, many deep models adopt count-likelihood objectives; we complement these with negative binomial and zero-inflated Poisson baselines in our appendix. These studies motivate a gray-box alternative that delivers interpretability and CPU efficiency while remaining competitive.

\paragraph{Physics-enhanced residual learning.} Recent theory formalizes why residualizing against a physics-based baseline can improve data efficiency and interpretability while preserving predictive accuracy \citep{perl_2025}. AMBIT instantiates this idea for OD prediction by treating a physical model as the explicit backbone and learning a structured residual.

\paragraph{OD-as-node and completion-based approaches.} Beyond grid-based spatiotemporal models, a growing line of work represents OD pairs (or stations) as nodes in multi-graphs and learns OD dynamics with graph neural networks and encoder--decoder architectures, including residual multi-graph convolutional models and inductive multi-graph representation learning \citep{sted_rmgc_2021,mgraphsage_2024}. Related work also explores OD prediction via convolutional architectures and matrix-style completion/generation, including channel-attentive CNNs for OD demand \citep{cas_cnn_2021}, probabilistic GNN generators for sparse OD flows \citep{prob_gnn_od_2024}, and gravity-guided neural decoders/generative models \citep{odgn_2023}. AMBIT is complementary to these approaches: rather than replacing the physical baseline, it uses the baseline as an explicit, interpretable backbone and learns a structured residual to preserve transparency and CPU-feasible deployment.

\paragraph{PPML gravity and fixed effects.} In econometrics, PPML is a standard estimator for gravity models with heteroskedastic counts, and fixed effects (including high-dimensional variants) are commonly used to control for unobserved heterogeneity \citep{santos_silva_2006,ppmlhdfe_2020,ppml_ipp_2020}. We treat FE-PPML as a robustness check on subsamples and briefly discuss incidental-parameter concerns in Appendix.

\paragraph{Monotone and interpretable tree models.} Recent work on monotone gradient-boosted trees and interpretable additive tree models quantifies the accuracy cost of enforcing monotonicity and proposes domain-informed monotone regularizers \citep{monotone_gami_2023,monotone_tradeoff_2025,monotone_regularizer_2025}. These studies motivate our monotone residual experiment as a practical interpretability stress test.

\section{Data and study area}
We use NYC TLC Yellow Taxi data from 2024-12 to 2025-11 (12 months). Trips are aggregated to hourly OD flows between taxi zones using TLC lookup tables and zone boundaries. POI features are sourced from OpenStreetMap and aggregated to taxi zones.

\paragraph{Scale and distribution.} The final dataset contains 15,752,628 OD-hour observations across 8,759 hours, 175 unique origin zones, and 232 unique destination zones (263 zones in the full TLC lookup). The flow distribution is heavy-tailed (median 1, 90th percentile 6, 99th percentile 17, max 153), highlighting sparsity typical of OD matrices.

\paragraph{Preprocessing.} Trips are filtered by duration (1--180 minutes) and distance (0.1--100 km). OD pairs are filtered using training-period totals only: we retain pairs with at least 200 total trips and then keep up to the top 30,000 pairs by volume; in our data this yields 7,415 pairs (about 10.7\% of the full $263\times263$ OD matrix). This reduces extreme sparsity while preserving major mobility patterns. We quantify the impact of filtering on sample size and performance in Appendix Tables~\ref{tab:filter_stats} and \ref{tab:filter_sensitivity}.

\paragraph{Distance definition.} Distances are computed as straight-line (Euclidean) distances between taxi-zone centroids after projecting zone geometries to UTM 18N (EPSG:32618, meter units), and then converted to kilometers. We treat this as an interpretable proxy for impedance; network travel-time distances are discussed as a limitation (Section~\ref{sec:limitations}).

\section{Features}
We use three interpretable feature groups:
\begin{itemize}
  \item \textbf{Spatial structure:} zone centroids, areas, and inter-zone distances (km).
  \item \textbf{POI density:} OSM POI counts aggregated by taxi zone (amenity/shop/office), with densities per square km and total POI density.
  \item \textbf{Temporal context:} hour-of-day, day-of-week, month, weekend flag, and hour-of-week.
\end{itemize}

\section{Methods}
\subsection{Physical baselines}
We evaluate a broad set of physical models:
\begin{itemize}
  \item \textbf{Unconstrained gravity} with different mass definitions (flow-based, POI-based).
  \item \textbf{PPML gravity}, which models flows as Poisson counts and is more robust for skewed count data.
  \item \textbf{Production- and attraction-constrained gravity} (power and exponential decay).
  \item \textbf{Competing destinations} (Fotheringham) with tuned competition parameter.
  \item \textbf{Intervening opportunities / OPS} type models.
  \item \textbf{Radiation} and a doubly constrained hourly gravity variant \citep{radiation_2012}.
  \item \textbf{Count-model baselines} (negative binomial, zero-inflated Poisson) on a subsample (Appendix Table~\ref{tab:count_baselines}).
\end{itemize}
We use standard shorthand in tables: DC = doubly constrained; OPS/IO = (intervening) opportunity-based formulations; ``hour'' denotes hour-of-day segmented gravity.

We tune decay and competition parameters on a validation split using explicit grids (e.g., $\beta,\gamma\in\{0.5,1.0,1.5,2.0,\ldots\}$; $\rho\in\{0.5,1.0,1.5,2.0\}$) and evaluate on the held-out test period.

\paragraph{PPML specification and margins.} Our PPML gravity regresses flows on $\log$ origin mass, $\log$ destination mass, and $\log$ distance (power-law decay). We report three variants: (i) PPML with $T>0$ (positive flows only), (ii) PPML with zero-augmented samples (``PPML all''), and (iii) PPML with origin, destination, and time fixed effects (PPML+FE), optionally including origin$\times$time and destination$\times$time interactions. Because full FE-PPML is memory-intensive, we estimate PPML+FE on a random row subsample (size reported in Appendix Table~\ref{tab:ppml_fe_meta}) and treat it as a strong baseline rather than a definitive full-matrix estimate (Appendix Table~\ref{tab:ppml_sensitivity}). Zero augmentation draws full OD matrices for sampled hours and retains zeros via downsampling to a fixed budget; we additionally vary the zero/positive ratio to test sensitivity (Appendix Table~\ref{tab:ppml_zero_aug}). Flow-based masses are computed from the training period only to avoid leakage. For constrained models, margins are calibrated using the full (unfiltered) OD matrices \emph{from the training period only}; a full-matrix evaluation using hour-of-week averages is reported separately (Appendix Table~\ref{tab:physical_fullmatrix}) and is not directly comparable to the truncated main task.

\subsection{Residual learning (AMBIT)}
AMBIT models residuals in log space:
\[
r = \log(1 + T) - \log(1 + T_{base}),
\]
\[
\hat{T} = \exp(\log(1 + T_{base}) + \hat{r}) - 1.
\]
Here $T_{base}$ is a physical baseline and $\hat{r}$ is learned by gradient-boosted trees. Baselines are fit on the training split, and their (log) predictions are used as residual features for all splits. This reconstruction preserves non-negativity while keeping residuals interpretable; we provide SHAP explanations for feature contributions \citep{lundberg_2017}. Residual models are evaluated against direct XGBoost prediction (a strong black-box comparator).
We also test a monotonicity-constrained variant that enforces non-increasing residual effects with respect to distance using XGBoost's monotone constraints; results are reported in Appendix Table~\ref{tab:mono}.
Unless otherwise stated, XGBoost uses a validated configuration with early stopping (Appendix Table~\ref{tab:xgb_sensitivity}) and fixed seeds for reproducibility.

\paragraph{Count-aware boosting objectives.} To align the tree baselines with count data, we additionally train XGBoost with Poisson and Tweedie objectives on the raw flow scale, both as direct predictors and as baseline-calibration models that include the physical prediction as a feature. Results are reported in Appendix Table~\ref{tab:count_objectives}.

\paragraph{Residual-feature ablation.} To quantify how much the residual learner relies on explicit baseline predictions, we report an ablation that removes the baseline prediction features while keeping the residual target unchanged (Appendix Table~\ref{tab:residual_base_ablation}).

\section{Experimental design}
We use time-based splits: train before 2025-08-01, validation 2025-08-01 to 2025-10-01, and test 2025-10-01 to 2025-12-01. Due to hardware constraints, we sample up to 3M training rows and 2M evaluation rows (seed 42); we support both random and stratified-by-flow sampling and report tail diagnostics by flow quantile (Appendix Table~\ref{tab:quantile_errors}). We also report multi-seed robustness (3 seeds; Appendix Table~\ref{tab:seed_robustness}). OD pairs are filtered to the top 30,000 by volume with at least 200 total trips \emph{using the training period only}; in practice this yields 7,415 pairs and defines the prediction task, biasing toward higher-volume pairs. For constrained models, margins are computed from the full unfiltered OD matrices in the training period to avoid truncation bias; this can create a calibration--evaluation mismatch on the truncated task because constrained models allocate flow to OD pairs outside the filtered set. We therefore also report a separate full-matrix evaluation (hour-of-week averages) in Appendix Table~\ref{tab:physical_fullmatrix}. Flow-based masses are computed from the training period only; POI masses are static.

Metrics include MAE, RMSE, $R^2$, CPC (common part of commuters), and sMAPE, defined as $\mathrm{sMAPE}=2|y-\hat{y}|/(|y|+|\hat{y}|)$ and reported on $[0,2]$. CPC is computed as $2\sum\min(y,\hat{y})/(\sum y+\sum\hat{y})$; we clip predictions to be non-negative before computing CPC and other count metrics. CPC is reported as a global aggregation over OD-hour observations; we additionally report hour-averaged CPC in Appendix Table~\ref{tab:cpc_hourly}. Spatial generalization is assessed by holding out 10\% of zones uniformly at random; all OD pairs involving those zones are removed from training and models are fit on the remaining zones. Flow-based masses for holdout zones are computed from training only (and thus are zero for withheld zones), while POI masses remain available for all zones. To separate model capability from mass availability, we additionally report a spatial-holdout variant with imputed flow-based masses (Appendix Table~\ref{tab:spatial_holdout}) and a leave-one-borough-out evaluation (Appendix Table~\ref{tab:borough_holdout}). We also run a small-sample impedance sensitivity experiment that replaces Euclidean distance with a travel-time proxy derived from raw trips (Appendix Table~\ref{tab:impedance_sensitivity}). XGBoost uses validation-based early stopping and we report a small hyperparameter sensitivity analysis (Appendix Table~\ref{tab:xgb_sensitivity}); constrained gravity models tune decay/competition parameters via explicit grids and runtime is summarized in Appendix Table~\ref{tab:runtime}.

\section{Results}
\subsection{Physical model audit}
Table~\ref{tab:physical} shows that PPML gravity variants are the strongest physical baselines; PPML+FE (estimated on a representative subsample) is consistently best, while PPML (all) and PPML ($T>0$) follow. Several constrained power-law variants and the hourly doubly constrained model achieve positive $R^2$ when calibrated with full margins, but they still lag PPML. Radiation and IO/OPS remain weak. This suggests that at hourly resolution, simple physical assumptions are brittle, and careful calibration is necessary to obtain usable constrained baselines.

\begin{table}[H]
\centering
\caption{Physical model audit (test set).}
\label{tab:physical}
\resizebox{\columnwidth}{!}{%
\begin{tabular}{lllll}
\toprule
model & mae & rmse & r2 & cpc \\
\midrule
Gravity (flow mass) & 1.913 & 3.993 & -0.114 & 0.499 \\
Gravity (POI mass) & 1.943 & 4.187 & -0.225 & 0.478 \\
Gravity (PPML, $T>0$) & 1.773 & 3.374 & 0.204 & 0.681 \\
Gravity (PPML, all) & 1.749 & 3.348 & 0.216 & 0.681 \\
Gravity (PPML + FE) & 1.685 & 3.183 & 0.292 & 0.696 \\
Radiation & 2.807 & 5.028 & -0.767 & 0.255 \\
DC Gravity (hourly) & 1.943 & 3.591 & 0.099 & 0.539 \\
Origin-constrained (power) & 1.924 & 3.546 & 0.121 & 0.548 \\
Competing destinations & 1.921 & 3.602 & 0.093 & 0.526 \\
OPS (opportunities) & 2.182 & 4.079 & -0.163 & 0.502 \\
\bottomrule
\end{tabular}
}
\end{table}

\paragraph{Key conclusion.} PPML gravity remains the most reliable physical baseline family on hourly OD data, with PPML+FE offering the strongest fit under standard gravity practice and PPML (all, zero-aug) improving robustness to zeros. Constrained variants benefit from full-margin calibration yet remain notably worse. Additional models, PPML sensitivity checks, and the full-matrix constrained evaluation are reported in Appendix Tables~\ref{tab:physical_full}, \ref{tab:ppml_sensitivity}, and \ref{tab:physical_fullmatrix}. We treat PPML+FE as a strong baseline estimated on a CPU-feasible subsample rather than a full-matrix definitive estimator.

\subsection{AMBIT vs black-box learning}
Table~\ref{tab:main} summarizes performance for the best residual configuration and direct XGBoost. On the main test split, time-segmented or DC-anchored residuals can slightly edge out the POI-anchored residual, while the POI anchor remains the most robust under spatial holdout (Appendix Tables~\ref{tab:residual_ablation} and \ref{tab:residual_ablation_spatial}). Overall, physics-grounded residuals approach the accuracy of a strong tree-based predictor while retaining interpretable structure, showing that interpretability can be preserved with minimal accuracy loss.
\paragraph{Uncertainty.} We report mean $\pm$ 95\% confidence intervals across seeds for headline models in Appendix Table~\ref{tab:seed_robustness}.
\paragraph{Residual-anchor ablation.} We provide a complete ablation over residual anchors (flow-based gravity, POI-based gravity, PPML, time-segmented gravity, and doubly constrained gravity) and their seed robustness and spatial-holdout behavior in Appendix Tables~\ref{tab:residual_ablation}--\ref{tab:residual_ablation_ci}.

\begin{table}[H]
\centering
\caption{AMBIT vs direct learning (test set).}
\label{tab:main}
\resizebox{\columnwidth}{!}{%
\begin{tabular}{lllll}
\toprule
model & mae & rmse & r2 & cpc \\
\midrule
XGB Direct & 1.103 & 2.209 & 0.659 & 0.784 \\
AMBIT (Residual + Gravity POI) & 1.104 & 2.218 & 0.656 & 0.783 \\
Gravity (PPML, all) & 1.750 & 3.340 & 0.220 & 0.682 \\
Gravity (PPML + FE) & 1.690 & 3.174 & 0.296 & 0.695 \\
Gravity (PPML, $T>0$) & 1.775 & 3.368 & 0.207 & 0.682 \\
\bottomrule
\end{tabular}
}
\end{table}

\paragraph{Deep baseline subset.} On a representative subset (CPU-feasible), we benchmark a lightweight GRU and a small Transformer with temporal inputs and static OD features. We use early stopping and a small hyperparameter sweep (Appendix Table~\ref{tab:deep_tuning}); final results are reported in Appendix Table~\ref{tab:deep_baselines}. In this subset, deep baselines remain below tree-based models, so we avoid claims of parity and keep deep comparisons limited to the subset.

\paragraph{Count-aware objectives.} Using Poisson and Tweedie objectives for XGBoost, with and without baseline-calibration features, does not materially change the qualitative ranking between tree models and physical baselines (Appendix Table~\ref{tab:count_objectives}).

\subsection{Spatial generalization}
Spatial holdout performance deteriorates for all models, confirming that cross-zone generalization is a hard problem. The AMBIT residual anchored in POI mass performs best among residual variants under spatial holdout, suggesting that physically meaningful mass definitions improve generalization.

\subsection{Interpretability and diagnostics}
\begin{figure}[t]
\centering
\includegraphics[width=\columnwidth]{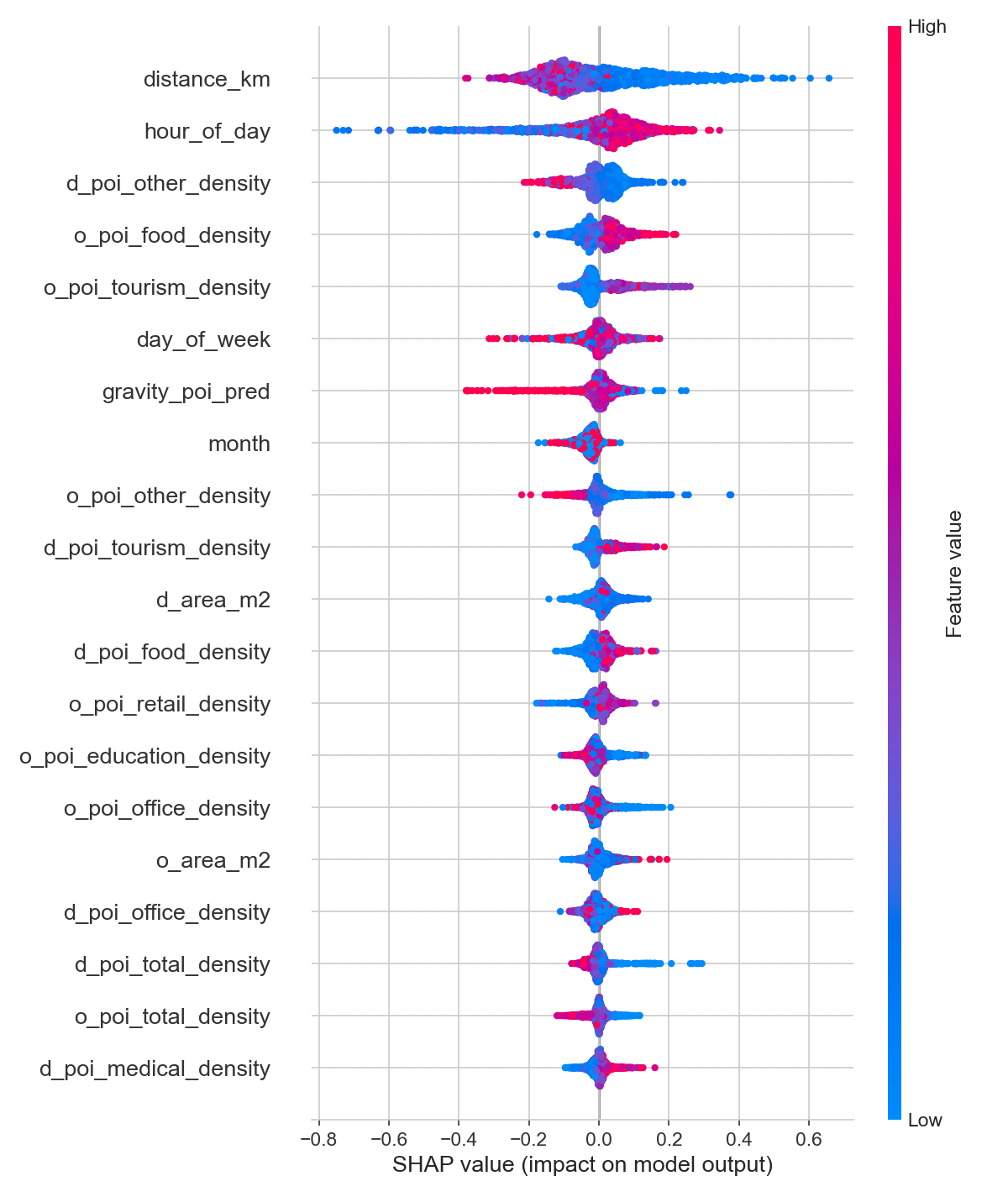}
\caption{SHAP summary for AMBIT residual (Gravity\_POI baseline).}
\label{fig:shap}
\end{figure}

Figure~\ref{fig:shap} summarizes SHAP attributions for the best residual model. The plot reports the distribution of per-feature contributions across test samples, with color indicating feature value; distance, POI density, and temporal features dominate, aligning with urban intuition. Positive distance contributions arise because the physical baseline already encodes decay and the residual corrects long-distance exceptions. We note that residual features include the baseline prediction itself, so SHAP attributions are distributed across correlated predictors; we therefore interpret rankings qualitatively rather than as exact causal effects.

\begin{figure}[t]
\centering
\begin{subfigure}{\columnwidth}
  \centering
  \includegraphics[width=\columnwidth]{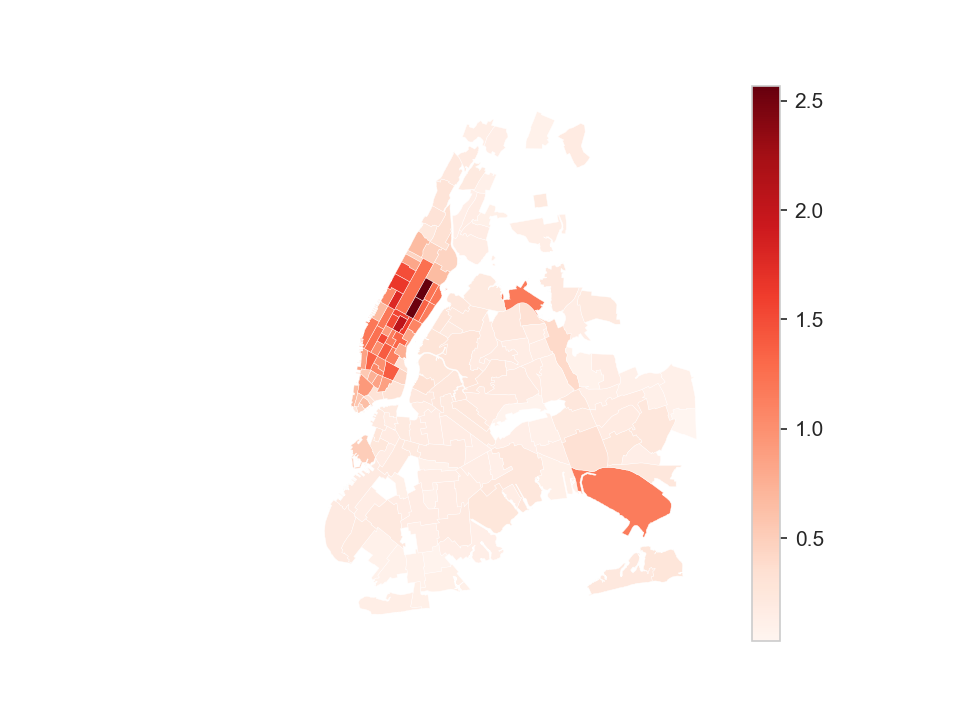}
  \caption{Absolute error (MAE by origin zone).}
\end{subfigure}
\vspace{0.5em}
\begin{subfigure}{\columnwidth}
  \centering
  \includegraphics[width=\columnwidth]{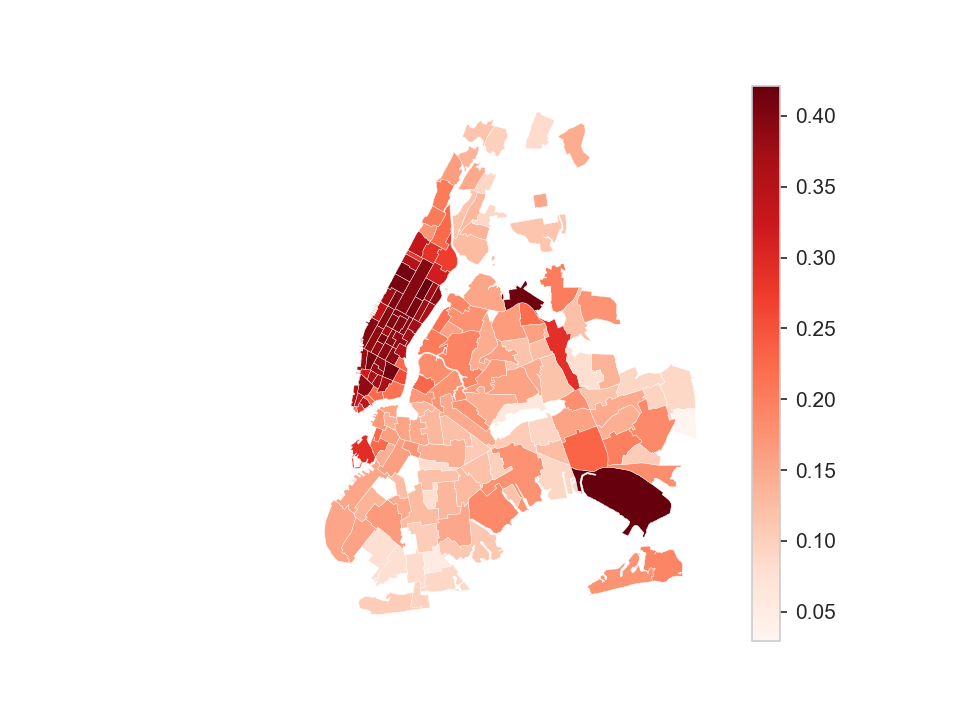}
  \caption{Relative error (sMAPE by origin zone).}
\end{subfigure}
\caption{Spatial error diagnostics for the AMBIT residual model.}
\label{fig:errors}
\end{figure}

Figure~\ref{fig:errors} maps spatial error by origin zone. We aggregate absolute errors to obtain zone-level MAE (left) and relative errors as sMAPE (right), highlighting spatial pockets where the residual model remains uncertain.

\begin{figure}[t]
\centering
\includegraphics[width=\columnwidth]{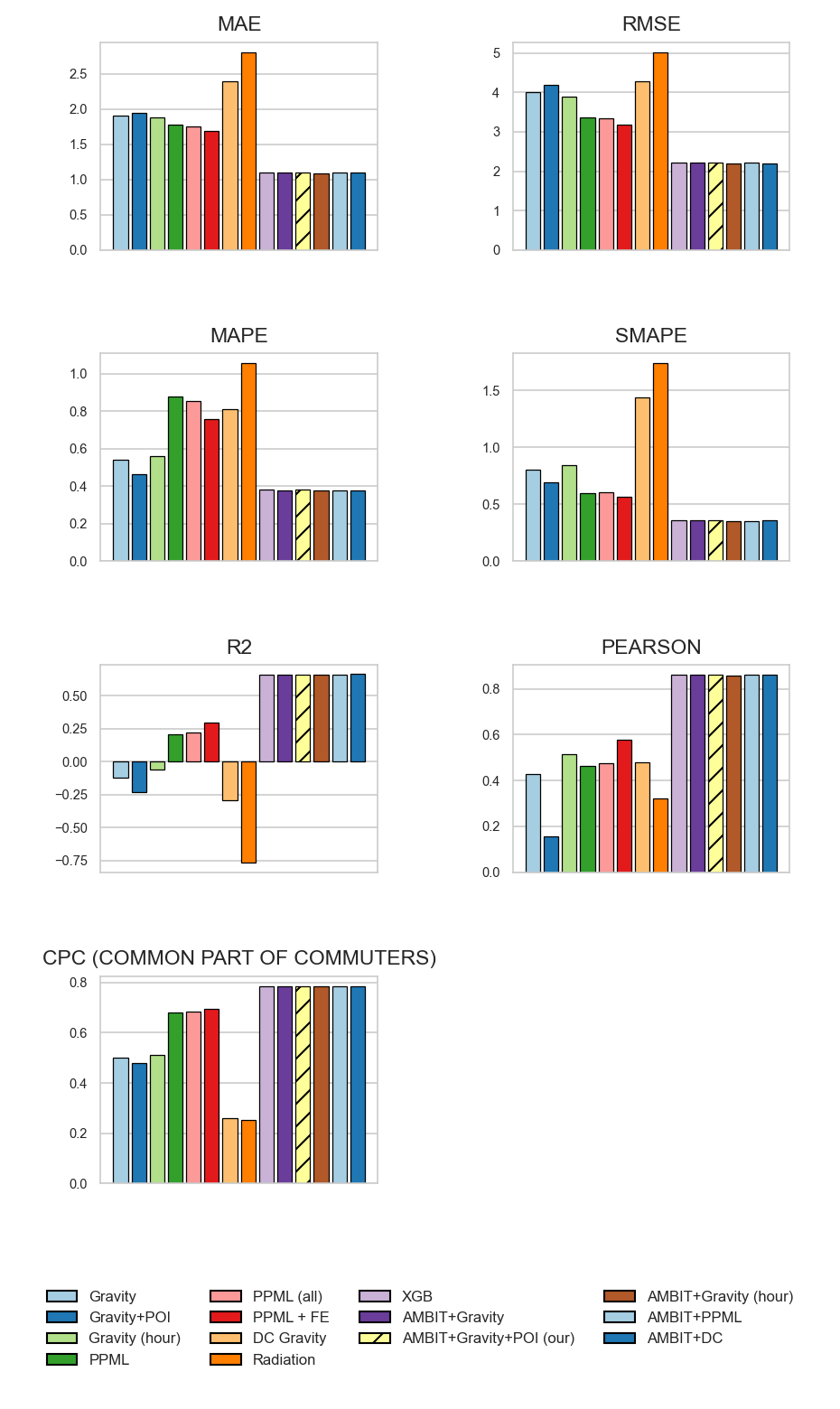}
\caption{Metric comparison across physical models and tree-based baselines (CPC = common part of commuters).}
\label{fig:metrics_bar}
\end{figure}

Figure~\ref{fig:metrics_bar} compares core metrics across physical baselines, residual variants, and the direct tree model on the test split to contextualize accuracy and interpretability trade-offs.

\begin{figure}[t]
\centering
\includegraphics[width=\columnwidth]{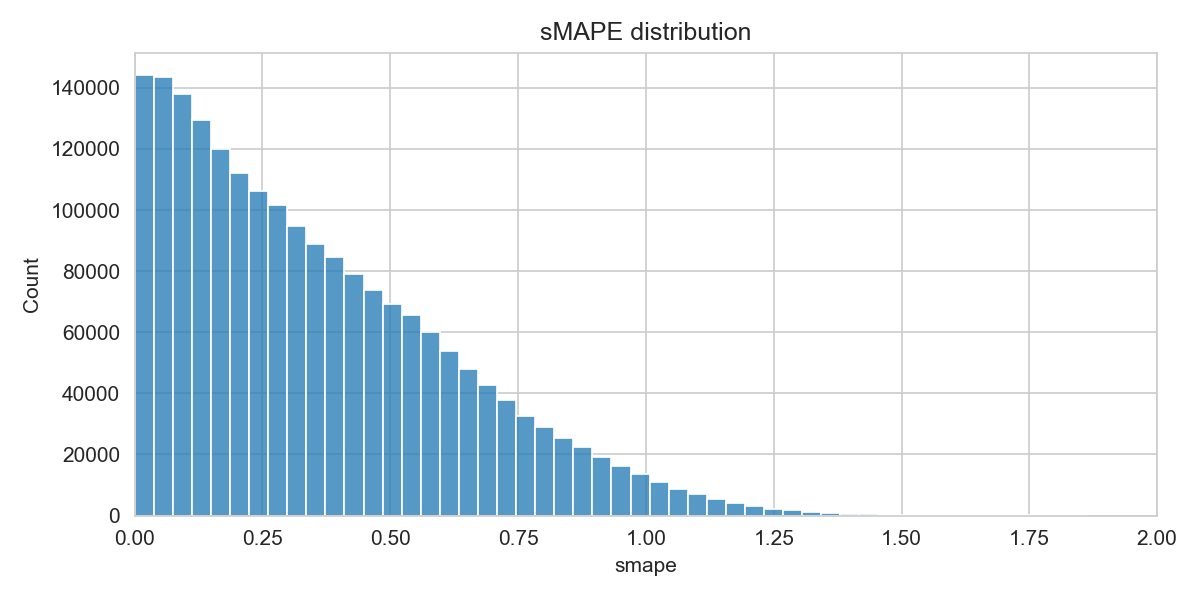}
\caption{sMAPE distribution across OD pairs (test set; sMAPE in $[0,2]$).}
\label{fig:error_hist}
\end{figure}

Figure~\ref{fig:error_hist} shows the distribution of sMAPE across OD pairs in the test set, illustrating the long-tailed error profile typical of sparse OD matrices.
We additionally report errors by flow quantile (Appendix Table~\ref{tab:quantile_errors}) to make tail behavior explicit.
We further report a residual-feature ablation that removes baseline prediction features to quantify how much the residual learner acts as a calibrator (Appendix Table~\ref{tab:residual_base_ablation}).

\section{Discussion}
\paragraph{Why physical models struggle at hourly resolution.} Classical models assume stable, slowly varying masses and smooth distance decay. Hourly taxi OD flows are sparse, heavy-tailed, and sensitive to local features. This mismatch causes most physical baselines to underperform, even when parameters are tuned; constrained variants improve with full-margin calibration but remain weaker than PPML. These findings are consistent with prior critiques of spatial interaction models under non-Gaussian, dynamic conditions \citep{sim_limitations_2020,gravity_migration_2022}.
In a small travel-time proxy experiment (18\% coverage), PPML and gravity improve modestly relative to Euclidean impedance, indicating that realistic impedance can help (Appendix Table~\ref{tab:impedance_sensitivity}).

\paragraph{Why PPML gravity works best.} The PPML-style formulation models counts directly, is robust to heteroskedasticity, and avoids log-transform bias. Under our main specification (log masses + log distance, no fixed effects), it outperforms other physical baselines and serves as the most reliable physical reference for residual learning; sensitivity checks with zeros and fixed effects are reported in Appendix Table~\ref{tab:ppml_sensitivity}.

\paragraph{Interpretability of residuals.} AMBIT retains a physical backbone while allowing residual corrections driven by POI density and time-of-day. In practice, this aligns with city planners' intuition: distance explains baseline flows, while urban amenities and temporal rhythms explain deviations.

\section{Limitations and future work}
\label{sec:limitations}
First, we use POI density as a proxy for attraction; true population/employment data may further improve physical baselines. Second, our distance impedance is straight-line centroid distance rather than network travel time; incorporating road-network or travel-time distances could materially strengthen physical baselines and is an important extension, though we provide a small travel-time proxy experiment in Appendix (Table~\ref{tab:impedance_sensitivity}). Third, the focus is on hourly taxi flows in NYC; generalization to other modes and cities is not guaranteed, though we include a same-city FHV sanity check in Appendix (Section~\ref{sec:fhv}). Fourth, spatial holdout results reveal that zone-level generalization remains challenging and deserves deeper structural modeling, even after imputing flow masses for held-out zones. Fifth, OD-pair filtering biases the task toward higher-volume pairs; we therefore quantify the effect of filtering (Appendix Tables~\ref{tab:filter_stats} and \ref{tab:filter_sensitivity}). Sixth, deep baselines are evaluated on a CPU-feasible subset rather than the full matrix; we therefore limit claims about parity with deep models to that subset. We now report multi-seed robustness and a monotonicity-constrained variant for distance in Appendix (Tables~\ref{tab:seed_robustness} and \ref{tab:mono}), as well as flow-quantile diagnostics (Table~\ref{tab:quantile_errors}). Finally, we add count-model baselines (negative binomial, zero-inflated Poisson) on a subsample in Appendix Table~\ref{tab:count_baselines} and leave richer exogenous variables (weather/events) for future work.

\section{Conclusion}
This study delivers a clear, empirically grounded message for urban analytics. Among classical physical models, PPML gravity is the most reliable baseline on hourly OD data; constrained variants improve with full-margin calibration but still lag. AMBIT then augments that physical baseline with interpretable tree-based residuals, achieving accuracy comparable to a strong black-box model while preserving interpretability and physical meaning. We provide a reproducible pipeline, comprehensive diagnostics, and transparent ablation results to guide future GIS mobility modeling.

\section*{Reproducibility}
All code, configurations, and experiment scripts will be released at \url{https://github.com/Icemap/ambit} upon acceptance. The raw taxi trip records are from the publicly available NYC TLC releases, and the POI features are derived from OpenStreetMap (ODbL; attribution required). The released repository will include instructions to reproduce preprocessing, training, and all paper tables/figures; no individual-level identifiers beyond the standard TLC fields are introduced by our pipeline, and our reported results operate on hourly aggregates at taxi-zone level.

\bibliographystyle{unsrt}
\bibliography{references}

\appendix
\section*{Appendix}
\addcontentsline{toc}{section}{Appendix}
\setcounter{table}{0}
\setcounter{figure}{0}
\renewcommand{\thetable}{A.\arabic{table}}
\renewcommand{\thefigure}{A.\arabic{figure}}
\section{Full physical model table}
This table reports all physical baseline variants on the main test set, including alternative mass and decay choices; it is intended as a comprehensive audit beyond the condensed main table.
\begin{table}[H]
\centering
\caption{Full physical model audit (test set).}
\label{tab:physical_full}
\scriptsize
\resizebox{\columnwidth}{!}{%
\begin{tabular}{llllll}
\toprule
model & mae & rmse & smape & r2 & cpc \\
\midrule
Gravity (flow mass) & 1.913 & 3.993 & 0.821 & -0.114 & 0.499 \\
Gravity (POI mass) & 1.943 & 4.187 & 0.701 & -0.225 & 0.478 \\
Gravity (PPML, $T>0$) & 1.773 & 3.374 & 0.606 & 0.204 & 0.681 \\
Gravity (PPML, all) & 1.749 & 3.348 & 0.610 & 0.216 & 0.681 \\
Gravity (PPML + FE) & 1.685 & 3.183 & 0.557 & 0.292 & 0.696 \\
Radiation & 2.807 & 5.028 & 1.736 & -0.767 & 0.255 \\
DC Gravity (hourly) & 1.943 & 3.591 & 1.061 & 0.099 & 0.539 \\
Origin-constrained (power) & 1.924 & 3.546 & 1.061 & 0.121 & 0.548 \\
Origin-constrained (power, POI) & 2.197 & 4.093 & 1.217 & -0.171 & 0.401 \\
Origin-constrained (exp) & 2.042 & 3.731 & 1.170 & 0.027 & 0.538 \\
Origin-constrained (exp, POI) & 2.141 & 3.948 & 1.195 & -0.089 & 0.487 \\
Destination-constrained (power) & 1.901 & 3.515 & 1.044 & 0.136 & 0.551 \\
Competing destinations & 1.921 & 3.602 & 1.036 & 0.093 & 0.526 \\
Competing destinations (POI) & 2.212 & 4.120 & 1.234 & -0.186 & 0.381 \\
OPS (opportunities) & 2.182 & 4.079 & 1.170 & -0.163 & 0.502 \\
Intervening opportunities (flow) & 2.469 & 5.017 & 1.269 & -0.759 & 0.439 \\
OPS (POI) & 2.270 & 4.147 & 1.242 & -0.202 & 0.445 \\
Intervening opportunities (POI) & 2.476 & 4.760 & 1.311 & -0.583 & 0.406 \\
\bottomrule
\end{tabular}
}
\end{table}

\section{Full-matrix constrained evaluation}
This table evaluates constrained models on full OD matrices averaged by hour-of-week; results are not directly comparable to the truncated main task but help assess margin-calibrated behavior.
\begin{table}[H]
\centering
\caption{Full-matrix constrained evaluation (hour-of-week averages).}
\label{tab:physical_fullmatrix}
\scriptsize
\resizebox{\columnwidth}{!}{%
\begin{tabular}{llllll}
\toprule
model & mae & rmse & smape & r2 & cpc \\
\midrule
DC Gravity (hourly) & 0.060 & 0.471 & 1.746 & 0.317 & 0.489 \\
Origin-constrained (power) & 0.058 & 0.452 & 1.755 & 0.370 & 0.504 \\
Origin-constrained (power, POI) & 0.077 & 0.548 & 1.821 & 0.074 & 0.342 \\
Origin-constrained (exp) & 0.060 & 0.535 & 1.462 & 0.118 & 0.485 \\
Origin-constrained (exp, POI) & 0.068 & 0.563 & 1.555 & 0.025 & 0.421 \\
Destination-constrained (power) & 0.059 & 0.447 & 1.787 & 0.384 & 0.502 \\
Competing destinations & 0.059 & 0.436 & 1.766 & 0.414 & 0.489 \\
Competing destinations (POI) & 0.079 & 0.552 & 1.821 & 0.062 & 0.326 \\
OPS (opportunities) & 0.063 & 0.610 & 1.731 & -0.147 & 0.459 \\
Intervening opportunities (flow) & 0.071 & 0.810 & 1.731 & -1.022 & 0.399 \\
OPS (POI) & 0.072 & 0.594 & 1.821 & -0.089 & 0.388 \\
Intervening opportunities (POI) & 0.075 & 0.740 & 1.823 & -0.688 & 0.356 \\
\bottomrule
\end{tabular}
}
\end{table}

\section{PPML sensitivity checks}
\label{sec:ppml_sensitivity}
We compare PPML variants that include zeros and high-dimensional fixed effects (on a subsample) to illustrate how specification choices affect performance. The FE-PPML row uses origin, destination, and time fixed effects with origin$\times$time and destination$\times$time interactions on the subsample.
\begin{table}[H]
\centering
\caption{PPML sensitivity checks (test set).}
\label{tab:ppml_sensitivity}
\resizebox{\columnwidth}{!}{%
\begin{tabular}{lllll}
\toprule
model & mae & rmse & r2 & cpc \\
\midrule
PPML ($T>0$) & 1.773 & 3.374 & 0.204 & 0.681 \\
PPML (all, zero-aug) & 1.749 & 3.348 & 0.216 & 0.681 \\
PPML + FE & 1.685 & 3.183 & 0.292 & 0.696 \\
\bottomrule
\end{tabular}
}
\end{table}

We additionally compare FE-PPML against an XGBoost model trained on a matching-size subsample (Appendix Table~\ref{tab:ppml_fe_fairness}).

\section{PPML augmentation and FE subsampling details}
\label{sec:ppml_details}
We report zero-to-positive ratios under PPML (all), including a sensitivity sweep over higher zero-retention settings, and summarize the PPML+FE subsample size and fixed-effect design complexity.
\begin{table}[H]
\centering
\caption{Zero-augmentation statistics for PPML (all).}
\label{tab:ppml_zero_aug}
\resizebox{\columnwidth}{!}{%
\begin{tabular}{llllll}
\toprule
Setting & Sampled hours & Rows & Zeros & Positives & Zero/pos ratio \\
\midrule
base & 200.000 & 3987403.000 & 969073.000 & 3018330.000 & 0.321 \\
ratio\_0.32 & 200.000 & 3954762.000 & 957925.000 & 2996837.000 & 0.320 \\
ratio\_1.0 & 200.000 & 3965913.000 & 969076.000 & 2996837.000 & 0.323 \\
ratio\_3.0 & 200.000 & 3965913.000 & 969076.000 & 2996837.000 & 0.323 \\
\bottomrule
\end{tabular}
}
\end{table}

\begin{table}[H]
\centering
\caption{PPML+FE subsample size and design complexity.}
\label{tab:ppml_fe_meta}
\resizebox{\columnwidth}{!}{%
\begin{tabular}{lllllll}
\toprule
Rows (pre) & Rows (post) & Origins & Dests & Hours & One-hot cats & Train time (s) \\
\midrule
3987403.000 & 100000.000 & 262.000 & 262.000 & 168.000 & 51670.000 & 4.107 \\
\bottomrule
\end{tabular}
}
\end{table}

Note that the origin/destination counts in Table~\ref{tab:ppml_fe_meta} reflect the sampled design matrix used for FE-PPML; because zero-augmentation draws full OD matrices, these counts can approach the full zone set even when the main task is filtered.

\section{Seed robustness}
\label{sec:seed_robustness}
We report mean performance across multiple random seeds for sampling and training to quantify stability of the main conclusions.
\begin{table}[H]
\centering
\caption{Multi-seed robustness (mean across seeds).}
\label{tab:seed_robustness}
\resizebox{\columnwidth}{!}{%
\begin{tabular}{lllll}
\toprule
Model & MAE (mean ± CI) & RMSE (mean ± CI) & R2 (mean ± CI) & CPC (mean ± CI) \\
\midrule
Gravity (PPML, $T>0$) & 1.777 ± 0.001 & 3.375 ± 0.002 & 0.207 ± 0.001 & 0.682 ± 0.000 \\
XGB Direct & 1.103 ± 0.001 & 2.209 ± 0.008 & 0.660 ± 0.002 & 0.784 ± 0.000 \\
AMBIT (Residual + Gravity POI) & 1.104 ± 0.003 & 2.218 ± 0.016 & 0.657 ± 0.005 & 0.783 ± 0.001 \\
\bottomrule
\end{tabular}
}
\end{table}

\section{Residual-anchor ablation}
\label{sec:residual_ablation}
We report a full ablation over residual anchors (flow mass, POI mass, PPML, time-segmented gravity, and doubly constrained gravity), including robustness across seeds and spatial holdouts.
\begin{table}[H]
\centering
\caption{Residual-anchor ablation on the main test split.}
\label{tab:residual_ablation}
\resizebox{\columnwidth}{!}{%
\begin{tabular}{lllll}
\toprule
model & mae & rmse & r2 & cpc \\
\midrule
xgb residual gravity time & 1.091 & 2.204 & 0.660 & 0.786 \\
xgb residual gravity dc & 1.094 & 2.196 & 0.663 & 0.786 \\
xgb residual gravity ppml & 1.096 & 2.216 & 0.657 & 0.785 \\
xgb residual gravity flow & 1.097 & 2.212 & 0.658 & 0.785 \\
AMBIT (Residual + Gravity POI) & 1.104 & 2.218 & 0.656 & 0.783 \\
AMBIT (Gravity POI, no base feat) & 1.107 & 2.224 & 0.654 & 0.783 \\
\bottomrule
\end{tabular}
}
\end{table}

\begin{table}[H]
\centering
\caption{Residual-anchor ablation under spatial holdout.}
\label{tab:residual_ablation_spatial}
\resizebox{\columnwidth}{!}{%
\begin{tabular}{lllll}
\toprule
model & mae & rmse & r2 & cpc \\
\midrule
AMBIT (Residual + Gravity POI) & 1.468 & 3.108 & 0.214 & 0.678 \\
xgb residual gravity dc & 1.708 & 3.621 & -0.067 & 0.591 \\
xgb residual gravity ppml & 1.980 & 3.996 & -0.299 & 0.469 \\
xgb residual gravity ppml all & 1.980 & 3.996 & -0.299 & 0.469 \\
xgb residual gravity time & 2.047 & 4.011 & -0.310 & 0.444 \\
xgb residual gravity flow & 2.503 & 4.252 & -0.472 & 0.218 \\
\bottomrule
\end{tabular}
}
\end{table}

\begin{table}[H]
\centering
\caption{Residual-anchor robustness (mean across seeds).}
\label{tab:residual_ablation_ci}
\resizebox{\columnwidth}{!}{%
\begin{tabular}{lllll}
\toprule
Model & MAE (mean ± CI) & RMSE (mean ± CI) & R2 (mean ± CI) & CPC (mean ± CI) \\
\midrule
XGB Direct & 1.103 ± 0.001 & 2.209 ± 0.008 & 0.660 ± 0.002 & 0.784 ± 0.000 \\
xgb residual gravity dc & 1.094 ± 0.001 & 2.196 ± 0.011 & 0.664 ± 0.003 & 0.786 ± 0.000 \\
xgb residual gravity flow & 1.097 ± 0.001 & 2.209 ± 0.001 & 0.660 ± 0.001 & 0.785 ± 0.000 \\
AMBIT (Residual + Gravity POI) & 1.104 ± 0.003 & 2.218 ± 0.016 & 0.657 ± 0.005 & 0.783 ± 0.001 \\
xgb residual gravity ppml & 1.097 ± 0.001 & 2.217 ± 0.002 & 0.658 ± 0.001 & 0.785 ± 0.000 \\
xgb residual gravity time & 1.091 ± 0.001 & 2.198 ± 0.008 & 0.664 ± 0.002 & 0.786 ± 0.000 \\
\bottomrule
\end{tabular}
}
\end{table}

\section{Count-aware boosting objectives}
\label{sec:count_objectives}
We compare Poisson and Tweedie objectives for direct tree models and baseline-calibration variants.
\begin{table}[H]
\centering
\caption{Count-aware boosting objectives (test set).}
\label{tab:count_objectives}
\resizebox{\columnwidth}{!}{%
\begin{tabular}{lllll}
\toprule
model & mae & rmse & r2 & cpc \\
\midrule
XGB Direct & 1.103 & 2.209 & 0.659 & 0.784 \\
XGB (Poisson) & 1.105 & 2.115 & 0.687 & 0.793 \\
XGB (Tweedie) & 1.080 & 2.082 & 0.697 & 0.797 \\
XGB (Poisson + Gravity POI) & 1.104 & 2.114 & 0.688 & 0.793 \\
XGB (Tweedie + Gravity POI) & 1.080 & 2.090 & 0.695 & 0.797 \\
\bottomrule
\end{tabular}
}
\end{table}

\section{Residual-feature ablation}
\label{sec:residual_base_ablation}
We ablate baseline prediction features in the residual learner to quantify calibration dependence.
\begin{table}[H]
\centering
\caption{Residual ablation: removing baseline prediction features.}
\label{tab:residual_base_ablation}
\resizebox{\columnwidth}{!}{%
\begin{tabular}{lllll}
\toprule
model & mae & rmse & r2 & cpc \\
\midrule
AMBIT (Residual + Gravity POI) & 1.104 & 2.218 & 0.656 & 0.783 \\
AMBIT (Gravity POI, no base feat) & 1.107 & 2.224 & 0.654 & 0.783 \\
\bottomrule
\end{tabular}
}
\end{table}

\section{FE-PPML fairness check}
\label{sec:ppml_fe_fairness}
We compare FE-PPML against an XGBoost model trained on a matching-size subsample.
\begin{table}[H]
\centering
\caption{FE-PPML subsample fairness check (test set).}
\label{tab:ppml_fe_fairness}
\resizebox{\columnwidth}{!}{%
\begin{tabular}{lllll}
\toprule
model & mae & rmse & r2 & cpc \\
\midrule
XGB Direct & 1.103 & 2.209 & 0.659 & 0.784 \\
XGB (FE-sample size) & 1.132 & 2.268 & 0.640 & 0.778 \\
Gravity (PPML + FE) & 1.690 & 3.174 & 0.296 & 0.695 \\
\bottomrule
\end{tabular}
}
\end{table}

\section{Impedance sensitivity (travel-time proxy)}
\label{sec:impedance_sensitivity}
We replace Euclidean distance with a travel-time proxy derived from raw trips on a small sample.
\begin{table}[H]
\centering
\caption{Impedance sensitivity (travel-time proxy).}
\label{tab:impedance_sensitivity}
\resizebox{\columnwidth}{!}{%
\begin{tabular}{llllll}
\toprule
Model & mae & rmse & r2 & cpc & Coverage \\
\midrule
Gravity (travel-time) & 1.900 & 3.940 & -0.069 & 0.506 & 0.182 \\
PPML (travel-time) & 1.680 & 3.264 & 0.266 & 0.698 & 0.182 \\
\bottomrule
\end{tabular}
}
\end{table}

\section{SHAP example explanations}
\label{sec:shap_examples}
\begin{figure}[H]
\centering
\begin{subfigure}{\columnwidth}
  \centering
  \includegraphics[width=\columnwidth]{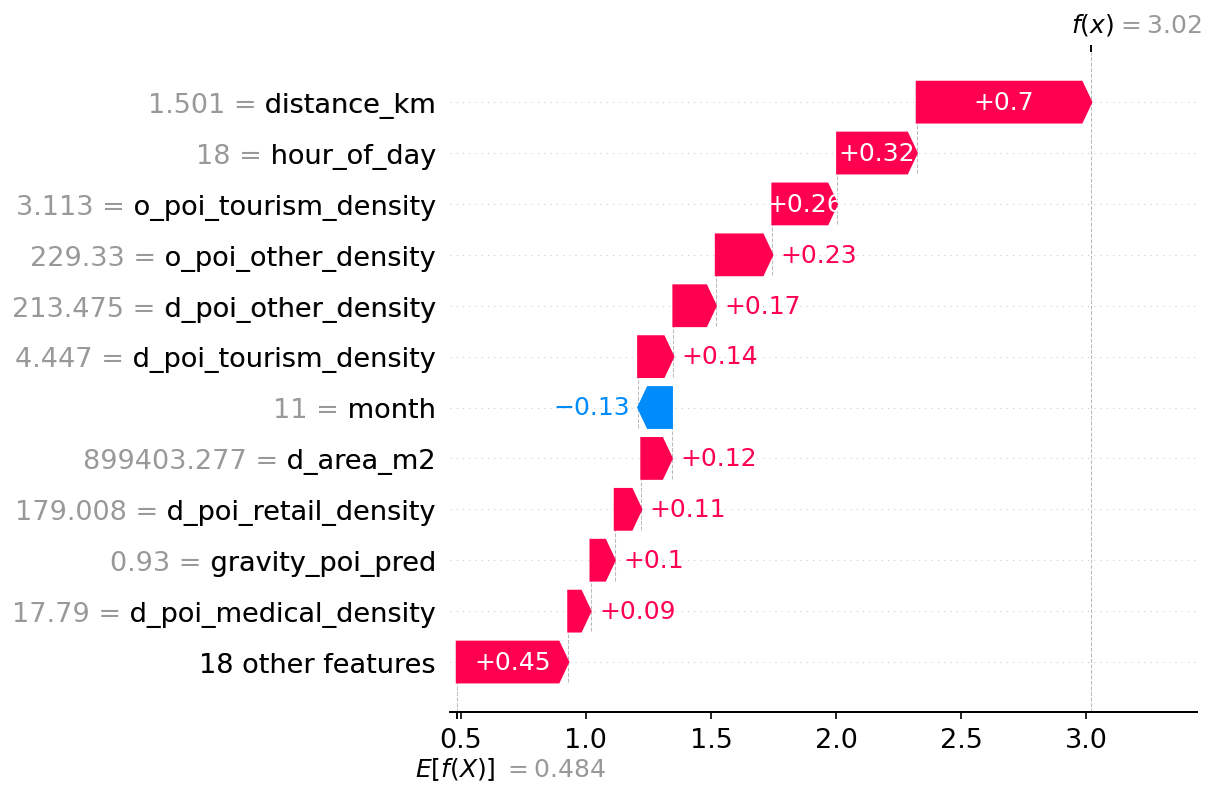}
  \caption{Example 1 (largest absolute error).}
\end{subfigure}
\vspace{0.5em}
\begin{subfigure}{\columnwidth}
  \centering
  \includegraphics[width=\columnwidth]{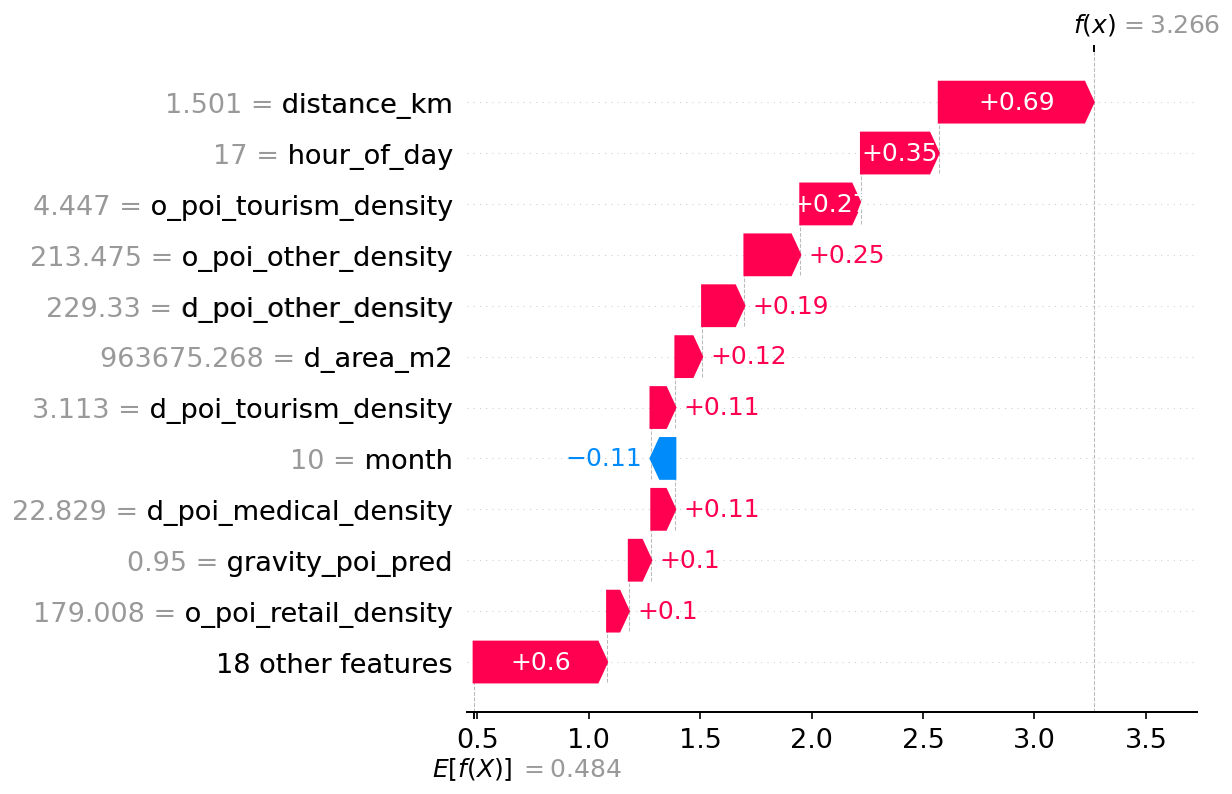}
  \caption{Example 2 (largest observed flow).}
\end{subfigure}
\vspace{0.5em}
\begin{subfigure}{\columnwidth}
  \centering
  \includegraphics[width=\columnwidth]{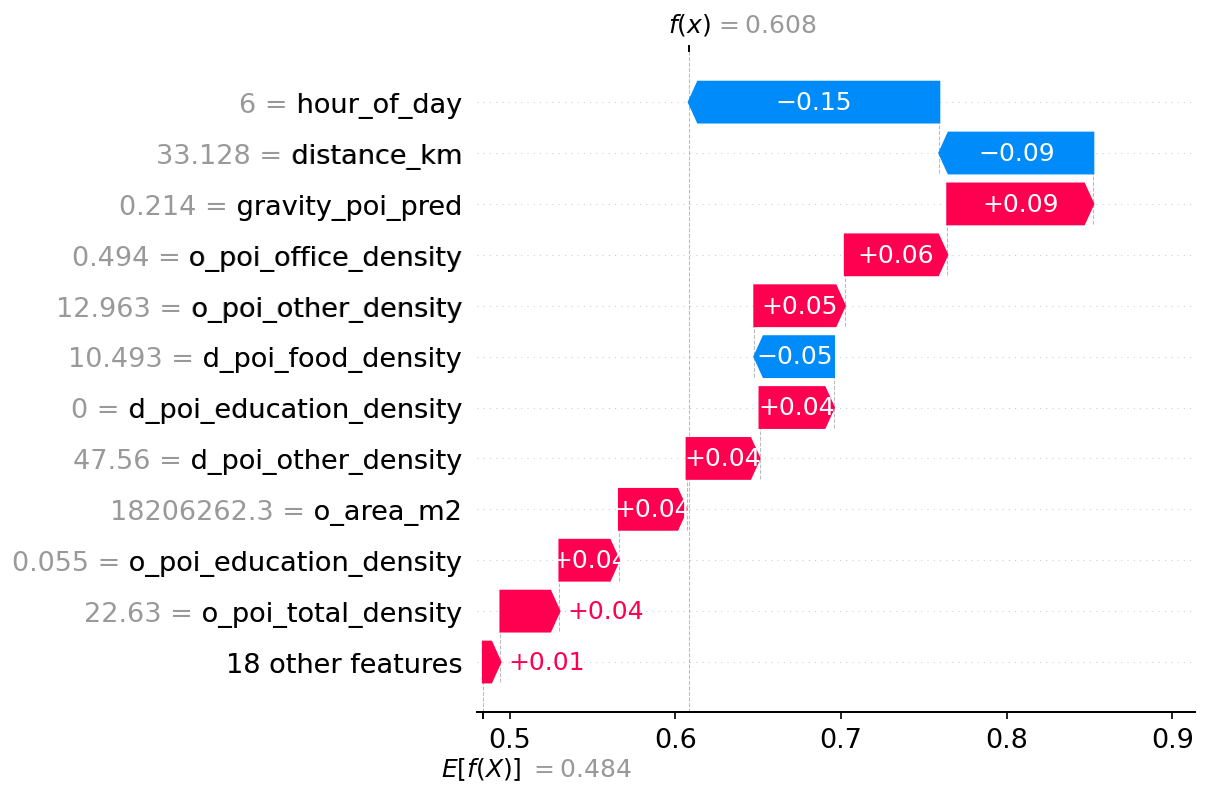}
  \caption{Example 3 (longest distance).}
\end{subfigure}
\caption{Example-level SHAP waterfall plots for the AMBIT residual model.}
\label{fig:shap_examples}
\end{figure}

We complement global SHAP summaries with example-level explanations (waterfall plots) for representative OD-hour cases, and report stability of SHAP feature rankings across early vs late test windows (Table~\ref{tab:shap_stability}).
Figure~\ref{fig:shap_examples} breaks down three representative OD-hour predictions into a baseline log-flow plus SHAP feature contributions (positive bars raise the prediction, negative bars reduce it). The examples are chosen to highlight distinct regimes---the largest absolute error, the largest observed flow, and the longest-distance trip---so the plot makes it clear how distance, POI density, temporal features, and the baseline prediction jointly drive each case.
\begin{table}[H]
\centering
\caption{Stability of SHAP feature ranking (early vs late test window).}
\label{tab:shap_stability}
\resizebox{\columnwidth}{!}{%
\begin{tabular}{lll}
\toprule
n/window & mid & Spearman $\rho$ \\
\midrule
2000.000 & 2025-10-31 14:00:00 & 0.996 \\
\bottomrule
\end{tabular}
}
\end{table}

\section{CPC aggregation sensitivity}
\label{sec:cpc_sensitivity}
We contrast global CPC with hour-averaged CPC to show that the ranking of models is robust to aggregation choices.
\begin{table}[H]
\centering
\caption{CPC sensitivity (global vs hour-averaged).}
\label{tab:cpc_hourly}
\resizebox{\columnwidth}{!}{%
\begin{tabular}{lll}
\toprule
Model & CPC (global) & CPC (hour-avg) \\
\midrule
Gravity (PPML, $T>0$) & 0.682 & 0.664 \\
XGB Direct & 0.784 & 0.796 \\
AMBIT (Residual + Gravity POI) & 0.783 & 0.796 \\
\bottomrule
\end{tabular}
}
\end{table}

\section{Monotonicity-constrained residuals}
\label{sec:mono}
This table shows the impact of enforcing non-increasing distance effects in residual trees, highlighting the trade-off between physical consistency and accuracy.
\begin{table}[H]
\centering
\caption{Residual models with monotonic distance constraints (test set).}
\label{tab:mono}
\resizebox{\columnwidth}{!}{%
\begin{tabular}{lllll}
\toprule
model & mae & rmse & r2 & cpc \\
\midrule
XGB Direct & 1.105 & 2.196 & 0.663 & 0.784 \\
xgb residual gravity flow & 1.099 & 2.220 & 0.655 & 0.784 \\
AMBIT (Residual + Gravity POI) & 1.102 & 2.200 & 0.662 & 0.784 \\
xgb residual gravity ppml & 1.099 & 2.214 & 0.657 & 0.784 \\
\bottomrule
\end{tabular}
}
\end{table}

\section{Spatial holdout with imputed masses}
\label{sec:spatial_impute}
We report a spatial-holdout variant where flow-based masses for held-out zones are imputed using borough-level averages from the training set to separate model capacity from mass availability.
\begin{table}[H]
\centering
\caption{Spatial holdout performance (test set).}
\label{tab:spatial_holdout}
\resizebox{\columnwidth}{!}{%
\begin{tabular}{lllll}
\toprule
model & mae & rmse & r2 & cpc \\
\midrule
Gravity (flow mass) & 2.852 & 4.519 & -0.662 & 0.000 \\
Gravity (POI mass) & 1.959 & 3.951 & -0.270 & 0.487 \\
Gravity (PPML, $T>0$) & 2.814 & 4.492 & -0.642 & 0.026 \\
XGB Direct & 1.512 & 3.228 & 0.152 & 0.663 \\
AMBIT (Residual + Gravity POI) & 1.468 & 3.108 & 0.214 & 0.678 \\
gravity flow imputed & 2.175 & 4.051 & -0.336 & 0.385 \\
gravity ppml imputed & 1.759 & 3.498 & 0.004 & 0.628 \\
\bottomrule
\end{tabular}
}
\end{table}

\section{Leave-one-borough-out evaluation}
\label{sec:borough_holdout}
We further assess spatial generalization by holding out all OD pairs involving a single borough at a time.
\begin{table}[H]
\centering
\caption{Leave-one-borough-out evaluation.}
\label{tab:borough_holdout}
\scriptsize
\resizebox{\columnwidth}{!}{%
\begin{tabular}{llllll}
\toprule
model & borough & mae & rmse & r2 & cpc \\
\midrule
Gravity (flow mass) & Bronx & 1.125 & 1.190 & -8.425 & 0.000 \\
Gravity (POI mass) & Bronx & 0.738 & 0.855 & -3.866 & 0.518 \\
Gravity (PPML, $T>0$) & Bronx & 1.085 & 1.152 & -7.827 & 0.068 \\
XGB Direct & Bronx & 0.264 & 0.419 & -0.166 & 0.885 \\
AMBIT (Residual + Gravity POI) & Bronx & 0.265 & 0.423 & -0.190 & 0.885 \\
Gravity (flow mass) & Brooklyn & 1.270 & 1.462 & -3.070 & 0.000 \\
Gravity (POI mass) & Brooklyn & 0.493 & 0.902 & -0.550 & 0.768 \\
Gravity (PPML, $T>0$) & Brooklyn & 1.253 & 1.446 & -2.980 & 0.027 \\
XGB Direct & Brooklyn & 0.614 & 1.058 & -1.130 & 0.786 \\
AMBIT (Residual + Gravity POI) & Brooklyn & 0.631 & 1.073 & -1.190 & 0.781 \\
Gravity (flow mass) & EWR & 1.434 & 1.717 & -2.310 & 0.000 \\
Gravity (POI mass) & EWR & 1.061 & 1.422 & -1.270 & 0.415 \\
Gravity (PPML, $T>0$) & EWR & 1.382 & 1.672 & -2.142 & 0.069 \\
XGB Direct & EWR & 0.603 & 0.896 & 0.098 & 0.804 \\
AMBIT (Residual + Gravity POI) & EWR & 0.504 & 0.869 & 0.151 & 0.815 \\
Gravity (flow mass) & Manhattan & 2.916 & 4.785 & -0.591 & 0.000 \\
Gravity (POI mass) & Manhattan & 2.866 & 4.755 & -0.572 & 0.035 \\
Gravity (PPML, $T>0$) & Manhattan & 2.866 & 4.760 & -0.575 & 0.034 \\
XGB Direct & Manhattan & 1.893 & 4.191 & -0.221 & 0.529 \\
AMBIT (Residual + Gravity POI) & Manhattan & 1.901 & 4.201 & -0.227 & 0.525 \\
Gravity (flow mass) & Queens & 2.054 & 3.060 & -0.821 & 0.000 \\
Gravity (POI mass) & Queens & 1.698 & 2.852 & -0.582 & 0.301 \\
Gravity (PPML, $T>0$) & Queens & 2.039 & 3.048 & -0.806 & 0.015 \\
XGB Direct & Queens & 1.127 & 2.462 & -0.178 & 0.648 \\
AMBIT (Residual + Gravity POI) & Queens & 1.155 & 2.479 & -0.195 & 0.633 \\
Gravity (flow mass) & Staten Island & 1.123 & 1.175 & -10.362 & 0.000 \\
Gravity (POI mass) & Staten Island & 0.485 & 0.598 & -1.941 & 0.724 \\
Gravity (PPML, $T>0$) & Staten Island & 1.055 & 1.112 & -9.161 & 0.113 \\
XGB Direct & Staten Island & 0.521 & 0.577 & -1.739 & 0.803 \\
AMBIT (Residual + Gravity POI) & Staten Island & 0.500 & 0.563 & -1.604 & 0.809 \\
\bottomrule
\end{tabular}
}
\end{table}

\section{Quantile error diagnostics}
\label{sec:quantile_errors}
This table reports errors by flow quantile to expose tail behavior under different models.
\begin{table}[H]
\centering
\caption{Error by flow quantile (test set).}
\label{tab:quantile_errors}
\scriptsize
\resizebox{\columnwidth}{!}{%
\begin{tabular}{lllllll}
\toprule
Model & Bin & MAE & RMSE & sMAPE & R2 & CPC \\
\midrule
Gravity (PPML, $T>0$) & Q1 & 1.263 & 1.612 & 0.661 & -12.298 & 0.649 \\
Gravity (PPML, $T>0$) & Q2 & 0.892 & 1.169 & 0.277 & -4.793 & 0.867 \\
Gravity (PPML, $T>0$) & Q3 & 4.933 & 7.759 & 0.637 & -0.464 & 0.628 \\
XGB Direct & Q1 & 0.466 & 0.712 & 0.298 & -1.598 & 0.832 \\
XGB Direct & Q2 & 1.231 & 1.455 & 0.439 & -7.979 & 0.793 \\
XGB Direct & Q3 & 3.811 & 5.204 & 0.545 & 0.341 & 0.739 \\
AMBIT (Residual + Gravity POI) & Q1 & 0.466 & 0.712 & 0.298 & -1.595 & 0.832 \\
AMBIT (Residual + Gravity POI) & Q2 & 1.233 & 1.456 & 0.439 & -7.998 & 0.792 \\
AMBIT (Residual + Gravity POI) & Q3 & 3.812 & 5.229 & 0.546 & 0.335 & 0.739 \\
\bottomrule
\end{tabular}
}
\end{table}

\section{XGBoost sensitivity}
\label{sec:xgb_sensitivity}
We report a small hyperparameter sensitivity sweep for direct and residual XGBoost models.
\begin{table}[H]
\centering
\caption{XGBoost sensitivity (test set).}
\label{tab:xgb_sensitivity}
\scriptsize
\resizebox{\columnwidth}{!}{%
\begin{tabular}{llllll}
\toprule
Model & Setting & mae & rmse & r2 & cpc \\
\midrule
XGB Direct & base & 1.150 & 2.363 & 0.629 & 0.774 \\
AMBIT (Residual + Gravity POI) & base & 1.151 & 2.361 & 0.630 & 0.774 \\
XGB Direct & low\_lr & 1.119 & 2.257 & 0.662 & 0.781 \\
AMBIT (Residual + Gravity POI) & low\_lr & 1.121 & 2.284 & 0.653 & 0.780 \\
XGB Direct & deep & 1.122 & 2.278 & 0.655 & 0.781 \\
AMBIT (Residual + Gravity POI) & deep & 1.123 & 2.284 & 0.653 & 0.781 \\
\bottomrule
\end{tabular}
}
\end{table}

\section{Filtering sensitivity}
\label{sec:filter_sensitivity}
We summarize OD-pair filtering statistics and performance sensitivity under stricter and unfiltered settings.
\begin{table}[H]
\centering
\caption{OD filtering statistics.}
\label{tab:filter_stats}
\resizebox{\columnwidth}{!}{%
\begin{tabular}{llll}
\toprule
Setting & Pairs & Total flow & Rows \\
\midrule
top30k & 7415.000 & 43681938.000 & 15752628.000 \\
top5k & 5000.000 & 42537088.000 & 14691186.000 \\
\bottomrule
\end{tabular}
}
\end{table}

\begin{table}[H]
\centering
\caption{Sensitivity to OD-pair filtering (test set).}
\label{tab:filter_sensitivity}
\scriptsize
\resizebox{\columnwidth}{!}{%
\begin{tabular}{llllll}
\toprule
Setting & Model & mae & rmse & r2 & cpc \\
\midrule
top30k & Gravity (PPML, $T>0$) & 1.777 & 3.373 & 0.207 & 0.682 \\
top30k & XGB Direct & 1.104 & 2.211 & 0.659 & 0.784 \\
top30k & AMBIT (Residual + Gravity POI) & 1.108 & 2.239 & 0.651 & 0.783 \\
top5k & Gravity (PPML, $T>0$) & 1.871 & 3.486 & 0.203 & 0.680 \\
top5k & XGB Direct & 1.174 & 2.284 & 0.658 & 0.780 \\
top5k & AMBIT (Residual + Gravity POI) & 1.179 & 2.307 & 0.651 & 0.779 \\
\bottomrule
\end{tabular}
}
\end{table}

\section{Deep baselines (subset)}
\label{sec:deep_baselines}
We evaluate lightweight GRU and Transformer baselines on a representative subset due to CPU constraints.
\begin{table}[H]
\centering
\caption{Deep baselines on a representative subset.}
\label{tab:deep_baselines}
\resizebox{\columnwidth}{!}{%
\begin{tabular}{lllll}
\toprule
model & mae & rmse & r2 & cpc \\
\midrule
Deep GRU (subset) & 1.210 & 2.773 & -0.015 & 0.671 \\
Deep Transformer (subset) & 1.229 & 2.768 & -0.012 & 0.670 \\
\bottomrule
\end{tabular}
}
\end{table}

\begin{table}[H]
\centering
\caption{Deep baseline tuning (subset).}
\label{tab:deep_tuning}
\scriptsize
\resizebox{\columnwidth}{!}{%
\begin{tabular}{lllllll}
\toprule
Model & Setting & Val loss & RMSE & MAE & R2 & CPC \\
\midrule
Deep GRU & base & 0.183 & 2.770 & 1.222 & -0.013 & 0.671 \\
Deep GRU & wide & 0.182 & 2.773 & 1.210 & -0.015 & 0.671 \\
Deep GRU & low\_lr & 0.184 & 2.773 & 1.220 & -0.015 & 0.670 \\
Deep Transformer & base & 0.184 & 2.767 & 1.231 & -0.011 & 0.671 \\
Deep Transformer & wide & 0.184 & 2.774 & 1.225 & -0.016 & 0.669 \\
Deep Transformer & low\_lr & 0.184 & 2.768 & 1.229 & -0.012 & 0.670 \\
\bottomrule
\end{tabular}
}
\end{table}

\section{Runtime summary}
\label{sec:runtime}
We report approximate training and inference times for key models on the main task.
\begin{table}[H]
\centering
\caption{Runtime summary on the main task.}
\label{tab:runtime}
\resizebox{\columnwidth}{!}{%
\begin{tabular}{lll}
\toprule
Model & Train time (s) & Pred time (s) \\
\midrule
Gravity (flow mass) & 2.377 & 0.072 \\
Gravity (PPML, $T>0$) & 2.539 & 0.079 \\
Gravity (PPML, all) & 0.676 & 0.076 \\
Gravity (PPML + FE) & 4.107 & 1.558 \\
XGB Direct & 31.106 & 1.993 \\
AMBIT (Residual + Gravity POI) & 34.268 & 2.020 \\
\bottomrule
\end{tabular}
}
\end{table}

\section{Hyperparameters and compute budget}
\label{sec:hyperparams}
We summarize key hyperparameters, subset sizes, and compute-related limits for reproducibility.
\begin{table}[H]
\centering
\caption{Key hyperparameters and compute budgets.}
\label{tab:hyperparams}
\scriptsize
\resizebox{\columnwidth}{!}{%
\begin{tabular}{p{0.20\linewidth} p{0.80\linewidth}}
\toprule
Component & Setting \\
\midrule
Sampling (main) & train max rows=3,000,000\newline val max rows=1,000,000\newline eval max rows=2,000,000\newline random sampling \\
\midrule
XGBoost & n\_estimators=500\newline max\_depth=8\newline lr=0.05\newline subsample=0.8\newline colsample\_bytree=0.8\newline early stopping=50\newline objective=reg:squarederror\newline tree\_method=hist \\
\midrule
PPML FE & include\_zeros=true\newline fe\_max\_rows=100,000\newline fe\_hour\_col=hour\_of\_week\newline FE includes O, D, T and O$\times$T/D$\times$T interactions \\
\midrule
Zero-aug PPML & zero\_sample\_hours=200\newline zero\_sample\_max\_rows=1,000,000 \\
\midrule
Deep GRU/\newline Transformer (subset) & seq\_len=24\newline top\_pairs=5,000\newline min\_pair\_total=200\newline max\_train/val/test=200k/50k/50k\newline batch=256\newline epochs=10\newline lr=1e-3\newline hidden=64\newline layers=2\newline dropout=0.1\newline heads=4\newline small tuning grid (Table~\ref{tab:deep_tuning}) \\
\bottomrule
\end{tabular}
}
\end{table}

\section{Count-model baselines}
\label{sec:count_baselines}
We include negative binomial and zero-inflated Poisson baselines on a subsample to test whether explicit count distributions improve over physical baselines.
\begin{table}[H]
\centering
\caption{Count-model baselines on a subsample.}
\label{tab:count_baselines}
\resizebox{\columnwidth}{!}{%
\begin{tabular}{lllll}
\toprule
model & mae & rmse & r2 & cpc \\
\midrule
Negative Binomial & 1.706 & 3.362 & 0.219 & 0.691 \\
Zero-inflated Poisson & 1.709 & 3.316 & 0.241 & 0.693 \\
\bottomrule
\end{tabular}
}
\end{table}

\begin{figure}[H]
\centering
\begin{subfigure}{\columnwidth}
  \centering
  \includegraphics[width=\columnwidth]{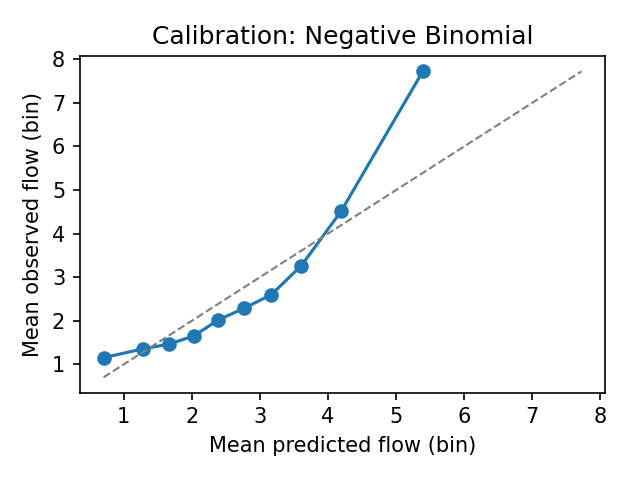}
  \caption{Negative binomial baseline.}
\end{subfigure}
\vspace{0.5em}
\begin{subfigure}{\columnwidth}
  \centering
  \includegraphics[width=\columnwidth]{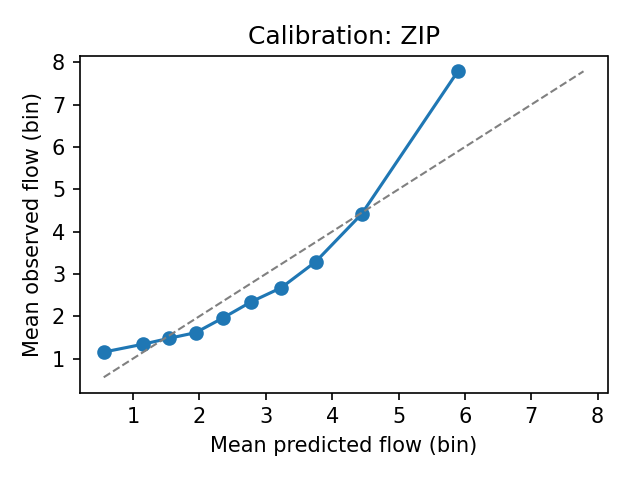}
  \caption{Zero-inflated Poisson baseline.}
\end{subfigure}
\caption{Calibration plots (decile-binned mean predicted vs mean observed) for count-model baselines on the test subsample.}
\label{fig:count_calibration}
\end{figure}

\section{Cross-mode sanity check (FHV)}
\label{sec:fhv}
This table reports the same core models on a small FHV sample to assess whether the qualitative findings transfer to another NYC mode.
\begin{table}[H]
\centering
\caption{Cross-mode sanity check on NYC FHV (test set).}
\label{tab:fhv_summary}
\resizebox{\columnwidth}{!}{%
\begin{tabular}{lllll}
\toprule
model & mae & rmse & r2 & cpc \\
\midrule
XGB Direct & 1.393 & 2.657 & 0.684 & 0.784 \\
AMBIT (Residual + Gravity POI) & 1.396 & 2.665 & 0.682 & 0.783 \\
Gravity (PPML, $T>0$) & 2.456 & 4.477 & 0.103 & 0.647 \\
\bottomrule
\end{tabular}
}
\end{table}

\end{document}